\documentclass[conference]{IEEEtran}
\IEEEoverridecommandlockouts

\usepackage{cite}
\usepackage{amsmath,amssymb,amsfonts}
\usepackage{algorithmic}
\usepackage{graphicx}
\usepackage{textcomp}
\usepackage{xcolor}
\usepackage{booktabs}
\usepackage{comment}
\def\BibTeX{{\rm B\kern-.05em{\sc i\kern-.025em b}\kern-.08em
    T\kern-.1667em\lower.7ex\hbox{E}\kern-.125emX}}
\begin{document}

\title{NLD-LLM: A systematic framework for evaluating small language transformer models on natural language description
 \\
{\footnotesize }
\thanks{}
}
\author{
Hamed Jelodar, Mohammad Meymani, Parisa Hamedi, Tochukwu Emmanuel Nwankwo,\\
Samita Bai, Roozbeh Razavi-Far, and Ali A. Ghorbani \\
\textit{Canadian Institute for Cybersecurity},\\
\textit{Faculty of Computer Science}, \\
\textit{University of New Brunswick} \\
Fredericton, Canada \\
\{h.jelodar, mohammad.meymani79, parisa.hamedi, tochukwu.nwankwo,\\
samita.bai, roozbeh.razavi-far, ghorbani\}@unb.ca
}
\maketitle

\begin{abstract}
Natural Language Description (NLD) is a Natural Language Processing (NLP) task that requires models to generate structured and meaningful outputs from natural language inputs. In this work, we propose NLD-LLM, a systematic NLP framework to evaluate the performance of language models to generate accurate and concise source code descriptions. This framework incorporates a diverse set of transformer models, including Qwen, DeepSeek, Phi, LLaMA, and Mistral, spanning various sizes, architectures, and training approaches. Central to NLD-LLM is a comprehensive prompt design strategy that includes standardized formatting, clear task guidance, and NLD prompting, ensuring fair and consistent evaluation. Additionally, we apply an iterative refinement process to improve output's quality and assess the model's adaptability. Using semantic and structural metrics, our analysis demonstrates that prompt engineering significantly impacts the effectiveness of the model such that smaller models often performing competitively when supported by well-crafted prompts.
\end{abstract}

\begin{IEEEkeywords}
Natural Language Description (NLD), Large Language Models, Prompt Engineering, Semantic and Structural Metrics
\end{IEEEkeywords}

\section{Introduction}
The ability to generate natural language descriptions from complex or structured data has become an increasingly important task in the field of Natural Language Processing (NLP). Such descriptions enable better human understanding and interaction with diverse types of data, ranging from technical netlists and programming models to visual scenes and regulatory documents. With rapid advancement of large language models (LLMs), there is a significant potential to use their generative capabilities for transforming structured inputs into coherent and contextually accurate natural language outputs \cite{shi2025natural, joshi2025natural, nelson2024enhancing, pavitha2024nl2code}.

Recent studies have explored various applications of LLMs for this purpose, including generating netlists \cite{1}, converting structured semantic graphs \cite{2}, formulating optimization problems \cite{3}, and describing complex traffic scenarios \cite{4}. These works demonstrate that task-specific frameworks combined with fine-tuning and prompt engineering can greatly enhance the performance of LLMs. However, the ability of these models to consistently produce high-quality natural language descriptions across different domains and input formats remains an open challenge. In this paper, we focus on evaluating the potential of state-of-the-art LLMs to generate natural language descriptions from diverse input types. Our goal is to systematically assess their strengths and limitations in translating structured or domain-specific data into clear, coherent, and contextually appropriate natural language \cite{jelodar2025large,jelodar2025large2} . By benchmarking several leading models and analyzing their outputs, we aim to provide insights into how well the current LLMs can perform this task and identify opportunities for future improvement.

\subsection{Research Contribution}

This research presents several key contributions to the field of Natural Language Description (NLD) tasks. Through systematic evaluation and experimentation, we explore how different language models perform in generating structured and meaningful descriptions from natural language inputs. Our main contributions are as follows:

\begin{itemize}

    \item We demonstrate that smaller models can perform competitively or even outperform larger models in specific generation tasks.
   
    \item We evaluate the performance of transformer models using various metrics such as semantic similarity and n-gram overlap.
    \item To the best of our knowledge, this is the first work that evaluates a range of pre-trained models for the NLD task based on small and large models aspects.
    \item We demonstrate that even small language models, such as Qwen \cite{bai2023qwen}, are capable of effectively generating NLDs.\\
\end{itemize}

The remainder of this paper is structured as follows. Section \ref{sec:related-work} reviews the related work in this area. Section \ref{sec:methodology} outlines our evaluation methodology, including dataset selection and model configurations. Section \ref{sec:experiments} presents the experimental results and analysis. Section \ref{sec:discussion} demonstrates our findings. Section \ref{sec:future-work} introduces directions for future research. Finally, in Section \ref{sec:conclusion}, we conclude our work.

\begin{figure*}[!htbp]
\centering
\includegraphics[width=0.96\linewidth]{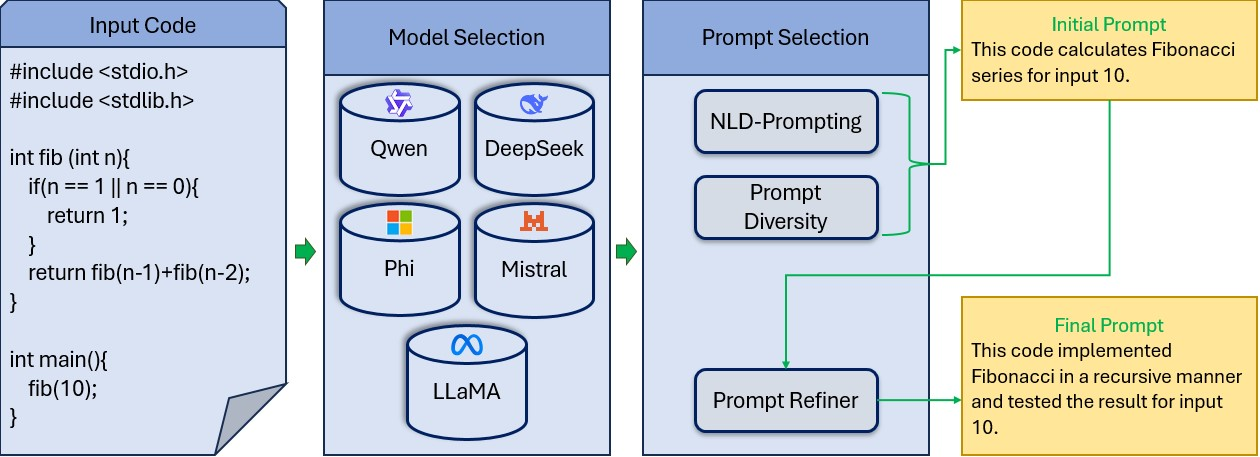}
\caption{A general view of the research model for NLD generation}
\label{fig:pipeline-overview}
\end{figure*}
\section{Related work}
\label{sec:related-work}
We outline the contributions of a few relevant research works here. In \cite{1}, the authors addressed the challenge of automatically generating hardware netlists from functional specifications written in natural language using LLMs. The authors developed specialized training and evaluation datasets and proposed LLM4NETLIST, a step-based framework with prompt engineering, fine-tuning, and feedback mechanisms.

In \cite{2}, the authors focused on improving how LLMs utilize structured representations such as Abstract Meaning Representation (AMR) graphs by proposing SR-LLM, a framework that converts these structures into natural language descriptions. Using models like T5 and LLaMA-2, they applied this approach through prompting and fine-tuning, demonstrating that natural language formats significantly boost model performance, whereas traditional code-like inputs can reduce it.

In \cite{3}, the authors concentrated on generating mixed-integer linear programming (MILP) models from natural languages using a three-step framework and fine-tuning an existing LLM (name not mentioned) for this task. They evaluated their approach against ChatGPT and Google Bard, showing that their method significantly outperformed both, achieving higher accuracy in identifying variables, constraints, and complete MILP formulations.

In \cite{4}, the authors introduced a framework that uses an off-the-shelf LLM (via the OpenAI API, presumably GPT-4 or GPT-3.5) to automatically construct detailed traffic scenes from natural language descriptions. Their pipeline decomposes text input using the LLM for prompt analysis, then retrieves suitable road layouts, plans agent behaviors, and ranks roads according to suitability, enabling realistic scene generation.

Building on these efforts, our work focuses on evaluating the potential of various LLMs and their ability to generate accurate and coherent natural language descriptions from complex inputs. We aim to assess how well current LLMs can translate structured or domain-specific data into natural language, highlighting strengths and limitations to inform future improvements.

\section{METHODOLOGY}
\label{sec:methodology}

\subsection{Model Selection}

For this study, we select a diverse set of language models representative of current state-of-the-art architectures in natural language processing. The models Qwen \cite{bai2023qwen}, DeepSeek \cite{liu2024deepseek}, Phi \cite{abdin2024phi}, LLaMA \cite{touvron2023llama}, and Mistral \cite{jiang2023mistral7b} were selected based on their different sizes, training strategies, and architectural designs. This diversity enables a thorough evaluation of how different model characteristics influence performance across various generation tasks (see Figure \ref{fig:pipeline-overview}).

Each model offers unique advantages: Qwen is known for its balanced trade-off between accuracy and efficiency; DeepSeek focuses on retrieval-augmented generation (RAG); Phi emphasizes scalability and speed; LLaMA has demonstrated strong generalization with moderate parameter sizes; and Mistral highlights efficient transformer design with competitive performance. By including these models, we aim to provide a comprehensive comparison that highlights the strengths and limitations of distinct NLP architectures, informing future research and deployment choices.

\begin{table*}[ht]
\centering
\caption{LLM Prompt Variations for Natural Language Description Generation for Source Code Description Use-case}
\begin{tabular}{|c|l|p{8cm}|p{4cm}|}
\hline
\textbf{\#} & \textbf{Prompt Style}           & \textbf{NLD Task Description }                                                                                                                                                                                                                             & \textbf{Notes}                                                                                  \\ \hline
1     & System Role-Based           & You are an AI assistant trained to analyze C and C++ code. Your task is to generate a natural language description that interprets the code’s logic, explains its purpose, or suggests relevant actions. Focus on producing a clear and purposeful explanation that reflects deep understanding. & Best for setting context in system prompts (e.g., OpenAI API system role).                 \\ \hline
2     & Zero-Shot Instruction       & Your task is to read the provided C/C++ code and produce a natural language description. The description should be purposeful, concise, and show an informed understanding of the code's behavior and intent.                                          & Direct and effective for one-off prompt completions.                                      \\ \hline
3     & Few-Shot Task               & For each C or C++ code snippet, generate a short explanation or recommendation in natural language. Your output should abstract away from syntax and emphasize what the code does and why it matters.                                                 & Ideal for few-shot prompting setups with examples.                                        \\ \hline
4     & System Message (Chat Setup) & You are a natural language description generator. Given a C or C++ function, produce a clear and insightful description or recommendation that captures the intent and behavior of the code.                                                            & Good for defining the LLM's role in conversational agents.                                \\ \hline
5     & Developer Tool Instruction  & Act as a smart code interpreter. When given a piece of C or C++ code, provide a clear, concise natural language summary that reflects what the code does and why it's written that way.                                                              & Useful for integration into IDE plugins or automated code documentation systems.          \\ \hline
6     & Concise One-Line Description & You are a Natural Language Descriptor (NLD) specialized in analyzing C and C++ source code. Your task is to generate a concise, one-line natural language description explaining what the code does.                                              & Useful for brief summaries and quick code understanding.                                 \\ \hline
\end{tabular}

\label{tab:llm-prompts}
\end{table*}

\begin{table*}[ht]
\centering
\caption{Key guidance points for designing effective NLD prompts for source code description use-case}
\begin{tabular}{|p{8.27cm}|p{8.27cm}|}
\hline
\textbf{NLD-Guidance for Task Clarity} & \textbf{NLD-Guidance for Prompt Quality} \\ \hline
Focus on the main purpose or effect of the code. Use no more than 50 words. & Avoid ambiguous wording. Ensure prompts are precise to reduce misunderstandings. \\ \hline
Use diverse prompt types. Test different aspects of source code comprehension. & Keep prompt length consistent. Maintain fairness and comparability across prompts. \\ \hline
Standardize prompt formats. Create uniformity to improve evaluation consistency. & Specify expected output style. Guide the model on how to format its response. \\ \hline
Reflect real-world scenarios in prompts. Make prompts relevant to practical coding tasks. & Use neutral phrasing to prevent bias. Avoid leading language that influences answers. \\ \hline
\end{tabular}

\label{tab:nld-prompt-guidance}
\end{table*}

\subsection{Prompt selection }
\begin{figure*}[h!]
\centering
\includegraphics[width=0.96\linewidth]{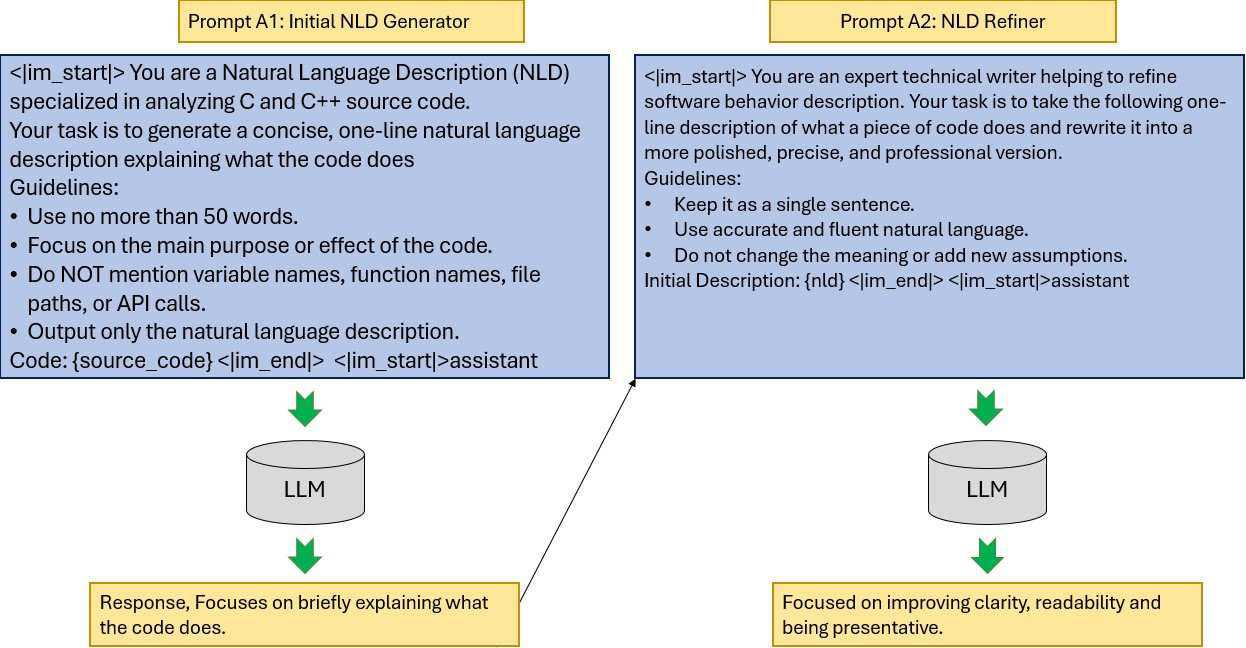}
\caption{Two-step Pipeline Prompt}
\label{fig:two-step-pipeline}
\end{figure*}

Prompt selection plays a critical role in evaluating language models, as the input prompts strongly influence the quality and relevance of the generated responses. In this study, we carefully curate a diverse set of prompts that cover various topics, formats, and complexity levels to ensure a comprehensive assessment. This includes both open-ended questions and more structured queries to test the models' abilities in reasoning, factual accuracy, and natural language understanding. By varying the prompt styles, we aim to capture the models' generalization capabilities and robustness across different real-world scenarios. Figure \ref{fig:two-step-pipeline} shows the prompt used for this framework.

Furthermore, we standardize prompt formatting to minimize ambiguity and bias, ensuring a fair comparison across all models. Each prompt is designed to elicit meaningful and contextually appropriate responses, facilitating reliable evaluation through quantitative metrics. 
 
\subsubsection{NLD-Prompting (Task definition)}
To better understand the framework, we focus on the source code description use case. For creating an NLD task, we identify different types of task prompts, as summarized in Table \ref{tab:llm-prompts}. These prompts vary in style and context to evaluate the models' ability to generate meaningful natural language descriptions under different conditions. Regarding Figure \ref{fig:two-step-pipeline}, we illustrate the use of the Concise One-Line Description approach.

\subsubsection{Prompt-NLD Guidance}
Prompt-NLD guidance plays a crucial role in designing effective natural language description tasks for source code understanding. It ensures that prompts are clear, consistent, and diverse enough to thoroughly evaluate a model’s comprehension and reasoning capabilities. Effective guidance can be broadly categorized into two aspects: task clarity and prompt quality.\\

Task clarity focuses on providing clear and structured directions that help the model understand exactly what is expected. This includes using diverse prompt types to cover multiple facets of code comprehension, standardizing prompt formats to ensure uniformity, and reflecting real-world scenarios to make the evaluation relevant to practical applications.\\

On the other hand, prompt quality emphasizes the formulation of prompts to minimize ambiguity and bias. This involves avoiding vague or leading language, maintaining consistent prompt lengths for fairness, specifying the expected style of output to guide the models' responses, and employing neutral phrasing to prevent influencing the model’s description.\\

Together, these guidelines help to create a balanced, robust, and fair evaluation framework that challenges language models to generate accurate, relevant, and meaningful natural language descriptions of source code. Table II summarizes the key guidance points for crafting prompts under these two complementary aspects.

\subsubsection{Prompt A1: Initial NLD Generator}
By considering the tasks and guidance outlined earlier, we prepare our initial prompt structure composing of two main sections: task description and guidance. The task description clearly defines the objective for generating natural language descriptions of source code, while the guidance provides detailed instructions to ensure clarity, consistency, and quality in the model’s responses.

\subsubsection{Prompt Standardization and NLD Refiner}

In the final step, we investigate the model’s behavior across different types of task descriptions and guidance to determine which prompts were the most suitable and meaningful. Through this evaluation, we refine and standardize the prompts to establish an optimal format for NLD generation that balances clarity, precision, and effectiveness. In Figure 2, we use the Concise One-Line Description prompt because it provides quick, clear, and focused explanations of code behavior, facilitating faster comprehension. Additionally, as shown on the right side of Figure 2, we incorporate a refiner strategy to generate more meaningful and polished outputs by iteratively improving the initial descriptions, ensuring higher quality and consistency in the generated natural language descriptions.

\section{Experiments}
\label{sec:experiments}
\subsection{ Model Configuration}
We consider LLMs to represent the performance ceiling in instruction-following and reasoning tasks. These models are designed with 7 to 8 billion parameters and offer strong general-purpose capabilities. In this study, we evaluate models such as \textit{Mistral-7B-Instruct} \cite{jiang2023mistral7b}  and \textit{Meta-Llama-3.1-8B-Instruct}, both of which are widely used in real-world applications due to their balanced trade-off between accuracy and scalability.

We also include small language models, generally ranging from under 2 billion parameters, to assess performance in resource-constrained scenarios. These models include \textit{Phi-4-mini-instruct (1.3B)}, \textit{DeepSeek-Coder-1.3B-Instruct} \cite{guo2024deepseek}, and \textit{Qwen2.5-Coder-1.5B-Instruct} \cite{hui2024qwen2}. These models are optimized for lightweight deployment and specific use cases such as code generation, mobile inference, or edge computing, offering practical alternatives to larger models.

All models are evaluated under consistent settings, using unified prompts, decoding parameters (temperature = 0.7, top-p = 0.9), and benchmark datasets. This ensures a fair comparison of model behavior across different scales and use cases, from general-purpose NLP to code-centric applications.

\subsection{Metrics}

To rigorously assess the performance of the language models, we employ standard LLM metrics. These metrics allow us to quantitatively compare model outputs and ensure consistency across different evaluation metrics:

\begin{itemize}
    \item \textbf{BLEU} \cite{papineni2002bleu}: Measures the precision of $n$-gram overlaps between a generated sentence and one or more reference sentences. BLEU is particularly effective in evaluating tasks where token-level accuracy is crucial.

    \item \textbf{ROUGE-L} \cite{lin2004rouge}: Focuses on the longest common subsequence between the system output and the reference, emphasizing recall. ROUGE is widely used in summarization tasks due to its sensitivity to sequence structure.

    \item \textbf{METEOR} \cite{banerjee2005meteor}: Incorporates synonym matching, stemming, and word order penalties to improve over BLEU. It offers a more balanced assessment by accounting for both precision and recall.

    \item \textbf{BERTScore} \cite{zhang2019bertscore}: Utilizes contextualized embeddings from pre-trained BERT models to compute semantic similarity between generated and reference texts. BERTScore is particularly useful when surface-level variation is high but semantic fidelity is maintained.

    \item \textbf{MAUVE} \cite{pillutla2021mauve}: Quantifies the distributional distance between human-written and machine-generated text using divergence measures. It evaluates the holistic naturalness and coherence of text, offering insights beyond sentence-level similarity.
\end{itemize}

\subsection{Human-Monitored Dataset for Natural Language Code Description}
To enable reliable evaluation using metrics such as BERTScore and METEOR, we first manually prepared and annotated 15100 code examples with high-quality natural language descriptions. These manually crafted references serve as gold standards for quantitative comparison against the generated descriptions. By calculating similarity scores, BERTScore to capture semantic alignment and METEOR to reflect exact matches and linguistic variations, we achieve a more comprehensive and nuanced assessment of the quality and the accuracy of the generated code descriptions. In future releases, our goal is to expand this dataset such that it includes several hundred examples across broader range of programming paradigms.




\subsection{Computing Settings}

All experiments were conducted using the Hugging Face \texttt{transformers} libraries. Models were authenticated and downloaded through a secured access token. Inference was performed in a Python environment configured to utilize GPU (NVIDIA H100).

\section{Results and Discussion}
\label{sec:discussion}

\subsection{Metric Correlation Analysis}
In Figure 3, the pair plot illustrates the relationships between different evaluation metrics across all tested models. BLEU scores remain low (0.00-0.05), reflecting its limited sensitivity to tasks such as code description where exact phrase matches are uncommon. ROUGE-L values (0.125-0.25) indicate moderate recall of reference content, while METEOR scores (0.150-0.300) demonstrate variability in handling synonyms and fluency. BERTScore shows the tightest clustering (0.84-0.89), confirming its reliability in assessing semantic similarity.

\subsection{Model-Wise Performance Breakdown}
The stacked area plot in Figure 4 provides a comparative view of how each metric contributes to overall model performance. Qwen dominates with the largest BERTScore segment, aligning with its superior semantic accuracy. LLaMA shows strength in BLEU and ROUGE-L but slightly trails in BERTScore, indicating a trade-off between exact matches and contextual understanding. Smaller models like DeepSeek and Phi exhibit substantially thinner stacks, consistent with their lower scores across all metrics.

BERTScore accounts for the majority of each model's stacked area, emphasizing its importance as an evaluation metric. Meanwhile, BLEU's marginal contribution reinforces its limited utility for this task. The visualization clearly differentiates holistic performers (Qwen) from models with narrower strengths (LLaMA in n-gram metrics). These insights help to guide model selection based on application priorities, whether optimizing for semantic fidelity (BERTScore) or specific aspects like code template generation (BLEU/ROUGE-L). The plot also surfaces opportunities for smaller models to improve through targeted fine-tuning on semantic alignment.

\subsection{Efficiency Considerations}

Although output quality is crucial, practical deployment requires attention to inference efficiency, especially in latency-sensitive or resource-constrained environments.

\begin{itemize}
    \item \textbf{Inference Speed:} Larger, more capable models such as Qwen may incur higher latency due to increased computational complexity.
    \item \textbf{Memory Footprint:} Models such as Phi and DeepSeek typically require less memory, making them suitable for edge or mobile deployment.
    \item \textbf{Operational Cost:} When cloud infrastructure is involved, inference cost scales with both time and hardware resources. Thus, while Qwen and LLaMA provide high accuracy, they may be less economical at scale.
\end{itemize}

Table 1 presents a comparative analysis of five language models evaluated on standard metrics for natural language generation: BLEU, ROUGE-L, METEOR, and BERTScore. Among these models, Qwen achieves the highest BERTScore (0.8853) and strong METEOR (0.2999) and ROUGE-L (0.2517) scores, indicating superior semantic similarity and linguistic quality in generated code descriptions.

LLaMA leads in BLEU score (0.0413), demonstrating better exact n-gram overlap, though its BERTScore (0.8752) and METEOR (0.2444) are slightly lower than Qwen’s. The other models, Mistral, Phi, and DeepSeek show lower scores across metrics, with DeepSeek particularly trailing in BLEU (0.0053) and ROUGE-L (0.1113), suggesting relatively weaker performance in matching the reference descriptions.

Overall, the results suggest that while Qwen excels in semantic fidelity and linguistic nuance, LLaMA performs well in more surface-level matching, and smaller models or code-focused models like DeepSeek and Phi show room for improvement in generating natural, high-quality descriptions.

\subsection{ Summary of Model Performance}

However, Qwen delivers the best overall performance across diverse quality metrics, making it the optimal choice when semantic fidelity and linguistic fluency are essential. LLaMA is a close second, with advantages in precision and a potentially more favorable efficiency profile. The decision between models should ultimately be guided by task-specific priorities and operational constraints. Table~\ref{tab:quant-metrics} summarizes the performance of five models in four widely used evaluation metrics.

\begin{table}[h!]
\centering
\caption{Quantitative Evaluation Results of Five Language Models.}
\begin{tabular}{lcccc}
\toprule \hline
\textbf{Model} & \textbf{BLEU} & \textbf{ROUGE-L} & \textbf{METEOR} & \textbf{BERTScore} \\
\midrule \hline
Qwen      & 0.0365 & 0.2517 & 0.2999 & 0.8853 \\
DeepSeek  & 0.0053 & 0.1113          & 0.1869          & 0.8478          \\
Phi       & 0.0108 & 0.1410          & 0.1558          & 0.8349          \\
LLaMA     & 0.0413 & 0.2239 & 0.2444 & 0.8752 \\
Mistral   & 0.0110 & 0.1432          & 0.2094          & 0.8565          \\
\bottomrule
\end{tabular}

\label{tab:quant-metrics}
\end{table}

\begin{figure}[h]
\centering
\includegraphics[width = \linewidth]{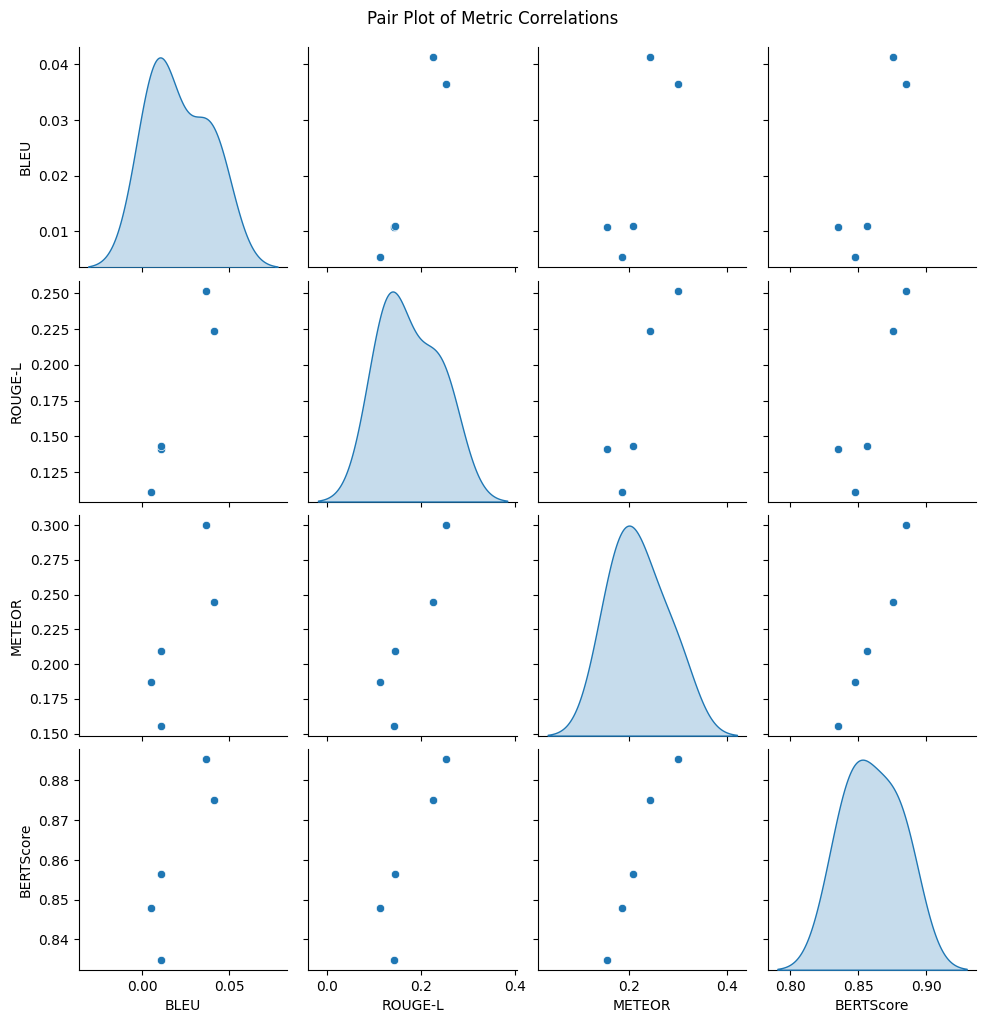}
\caption{A pair plot illustrating the relationships between different LLM evaluation metrics.}
\label{fig:method}
\end{figure}

\paragraph{Qwen} Qwen exhibits the most robust performance across the board. It leads in ROUGE-L, METEOR, and BERTScore, demonstrating superior coverage of reference content, sensitivity to synonymy and paraphrasing, and strong semantic alignment. Its BLEU score, while not the highest, remains competitive, suggesting that it balances exact match precision with deeper contextual understanding.

\paragraph{LLaMA} LLaMA achieves the highest BLEU score, indicating strong n-gram precision and surface-form overlap. It ranks second in METEOR and BERTScore, making it a close competitor to Qwen, particularly in tasks where precise wording matters.

\paragraph{Mistral and Phi} These mid-tier models show similar patterns, with Mistral marginally outperforming Phi on all metrics. Their relatively lower BLEU and ROUGE-L scores may suggest either more diverse generations or weaker task grounding, depending on the generation context.

\paragraph{DeepSeek} DeepSeek lags significantly in all metrics. This consistent underperformance may be attributed to suboptimal fine-tuning, reduced model capacity, or domain mismatch.


\begin{figure}[h]
\centering
\includegraphics[width=\linewidth]{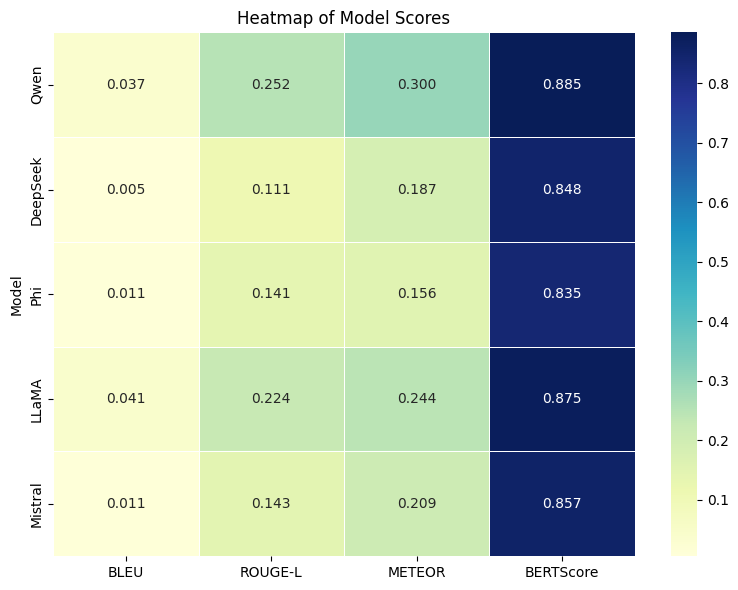}
\caption{Heatmap indicating the model scores}
\label{fig:method}
\end{figure}

\begin{figure}[h]
\centering
\includegraphics[width=\linewidth]{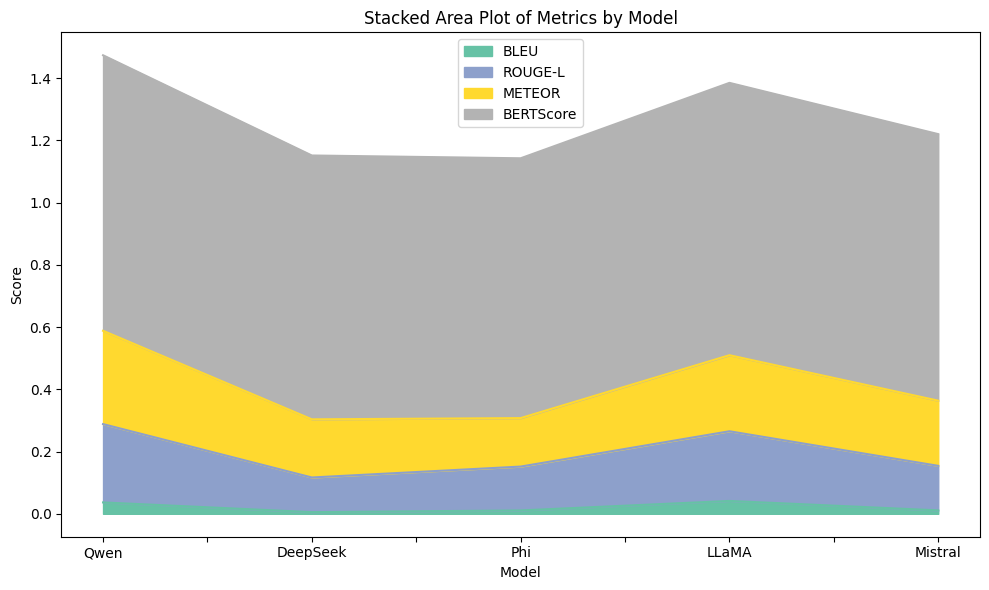}
\caption{A The stacked area plot for overall model performance}
\label{fig:method}
\end{figure}




\section{Discussion and Future Work}
\label{sec:future-work}
Based on our different experiments (especially Figure 5), we observe that Qwen and Mistral achieved similar accuracy compared to other methods. However, it is important to note that we used the 1.5 billion parameter version of the Qwen model, whereas the Mistral model had 7.5 billion parameters. This indicates that Qwen, although significantly smaller in size, can deliver comparable or even better accuracy than other larger models. 

\subsection{Model Size and Performance Comparison}
The experimental results indicates that the Qwen and Mistral models achieved similar levels of accuracy, outperforming several other baseline methods. In particular, the Qwen model that we used has 1.5 billion parameters, while the Mistral model has 7.5 billion parameters. Despite this significant difference in model size, both models performed nearly equally in terms of accuracy.

\subsection{Implications for Model Efficiency}
This finding suggests that Qwen, which is considerably smaller, demonstrates a higher parameter efficiency. In other words, a smaller model like Qwen can achieve a performance that is on par with much larger models. This has meaningful implications for resource-constrained environments, where deploying smaller, faster models with competitive accuracy can lead to better scalability, reduced latency, and lower computational costs.
\subsection{Limitations}
In this study, we only focused on NLD generation for C and C++ scripts, which decreases the breadth of the conclusion. In our future work, we plan to test other structures and programming languages such as Python and Java to increase the reliability of our framework results. Additionally, we note that our evaluation lacks frontier, large models such as Gemini \cite{team2023gemini} due to cost and access limitations. However, our goal is to focus on open-source, small-to-mid-scale models, which are more feasible for deployment in resource-constrained environments. Using closed-source models shifts our focus away from efficiency and reproducibility, which are central to our contributions.

\section{Conclusion}
\label{sec:conclusion}
This study systematically evaluated a range of LLMs, spanning both small- and large-scale architectures, to identify their effectiveness in NLD generation for the code. Our findings reveal that smaller, well-optimized models can rival or even surpass larger counterparts in generating concise and accurate text, challenging the common assumption that bigger models inherently perform better.

These results underscore the importance of selecting task-specific, resource-efficient models rather than defaulting to the largest available. Optimizing model choices based on the task and deployment constraints is critical for advancing scalable and practical LLM applications. Future work will focus on incorporating factors such as memory efficiency and output relevance to further refine model selection strategies.

While our framework uses conventional metrics, its novelty lies in integrating prompt standardization, iterative refinement, and systematic benchmarking into a unified pipeline. We also illustrate that small models can compete or surpass larger ones in semantic fidelity, offering valuable insights for researchers to balance cost, efficiency, and accuracy.

\bibliographystyle{ieeetr}
\bibliography{main}

\end{document}